\documentclass[runningheads]{llncs}

\usepackage{eccv}

\usepackage{eccvabbrv}

\usepackage{graphicx}
\usepackage{booktabs}
\usepackage{comment}

\usepackage[accsupp]{axessibility}  

\usepackage[pagebackref,breaklinks,colorlinks]{hyperref}

\usepackage{orcidlink}

\usepackage{pifont}
\usepackage{multirow}
\newcommand\blfootnote[1]{%
  \begingroup
  \renewcommand\thefootnote{}\footnote{#1}%
  \addtocounter{footnote}{-1}%
  \endgroup
}

\begin{document}

\title{InterAct: Capture and Modelling of Realistic, Expressive and Interactive Activities between Two Persons in Daily Scenarios} 

\titlerunning{Abbreviated paper title}

\author{Yinghao Huang $^{1,2,*}$ \and
Leo Ho $^{1,*}$ \and
Dafei Qin$^{1}$ \and \\
Mingyi Shi$^{1}$ \and
Taku Komura$^{1}$
}

\institute{
    The University of Hong Kong \and
    Centre for Transformative Garment Production
}

\maketitle

\begin{figure}
    \centering
    \includegraphics[width=0.95\textwidth]{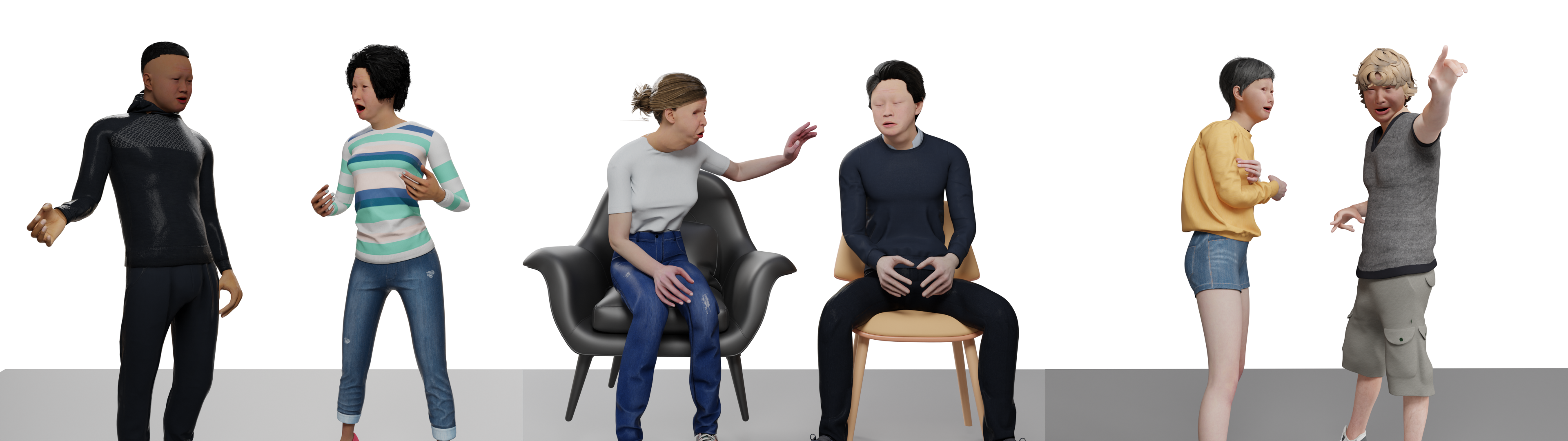}
    \caption{We capture full-body interactions between two persons in daily scenarios. Left: The fact that the man is a 1960s pop star surprises his neighbor. Middle: Boss(Female) comforts the employee(Male) who receives a sad phone call.  Right: Co-workers confess their romantic feelings.}
    \label{fig:teaser}
\end{figure}

\vspace{-8mm}

\begin{abstract}
We address the problem of accurate capture and expressive modelling of interactive behaviors happening between two persons in daily scenarios. Different from previous works which either only consider one person or focus on conversational gestures, we propose to simultaneously model the activities of two persons, and target objective-driven, dynamic, and coherent interactions which often span long duration. To this end, we capture a new dataset dubbed InterAct, which is composed of 241 motion sequences where two persons perform a realistic scenario over the whole sequence. The audios, body motions, and facial expressions of both persons are all captured in our dataset. We also demonstrate the first diffusion model based approach that directly estimates the interactive motions between two persons from their audios alone. All the data and code will be available at: \url{https://hku-cg.github.io/interact}. 
\end{abstract}
\blfootnote{$^*$ The first two authors contributed equally to this work. }

\section{Introduction}
The interaction between two persons during conversation is multi-modal: humans communicate not only through speech but also through facial expressions, body gestures and the space between the two. For example, even under the same speech, the response by the other person could be different according to the facial expressions, the body gesture of the speaker and the distance betwen the two. 

Previous works that produce facial motion and body gestures from speech do not fully consider such multi-modal nature of conversation.  Existing works mostly regress the speech data to the face \cite{fan2022faceformer, VOCA2019, karras2017audio} and body motion \cite{pang2023bodyformer, alexanderson2023listen} or condition the generative model to the speech contents and a prompt that describes the motion \cite{tevet2022human, xu2023inter, liu2022beat}. 
The works are mostly for the motion synthesis of a single agent. 
Although some recent works construct the full body and face motion from the speech \cite{ng2024audio2photoreal}, the setup is limited to subjects standing upright without much dynamic interactions between the two.  

In this paper, we propose the first method that jointly regresses realistic interactive body motions and detailed 3D face meshes for two persons from their speeches and their spatial relations. We target diverse and complex interaction happening between two persons in different scenarios, relationships and emotions, and let the characters move around in the space with rich facial expressions and full body motion, while keeping meaningful spatial relations during the conversation.

To train such a model, we prepare a large multi-modal motion capture dataset that includes the face, body and speech of two actors interacting with each other. The new dataset is dubbed InterAct. We prepare 241 scenarios which include a wide range of interactions where the actors express various emotions including anger, sadness, happiness, disgust, fear etc.  The data is prepared by first providing a short prompt of the scenario such as ``an interaction in a cafe where the customer orders and eats food but discovers a hair in the food and thus argue to the waiter'':  the actors then conduct a one minute act with rich emotional speech, facial expressions and body gestures. The dataset will be released upon publication of the paper. 

We show that our system can produce realistic interaction of the two agents from the speech, which includes both the face and body motion.  The results are evaluated by the FID score \cite{heusel2017gans} of the generated motion with respect to the test data and the variance of the motion distribution, and compared to other state-of-the-art motion synthesis approaches like LDA \cite{alexanderson2023listen}. Despite the simplicity of the scheme, our system shows the best performance in all criteria. 
Our system can be applied for the control of characters in films,  computer games and VR systems. 

\section{Related Works}

\subsection{Data-Driven Motion Synthesis from Speech}
Most classic works for audio-driven human motion generations are rule-based~\cite{levine2009real,neff2008gesture}: such systems can work well for constrained scenarios, but it is hard for them to generalize to more sophisticated cases in real life which are too complex to be clearly described by heuristic rules. In contrast, learning-based approaches target at automatically discovering the underlying mapping patterns between control signals and output motions:  we review such methods in the rest of this section.

Methods to regress the speech data to the face motion has been widely explored. 
Taylor et al.~\cite{taylor2017deep} regress the MFCC features to the face rig parameters by temporal convolution (TCN).  
Fan et al.~\cite{fan2022faceformer} make use of the transformer structure to learn the subtle correlation between the speech and the lips, which 
succeeds in correctly closing the mouth after the 'm' sound. 
VQ-based methods have been applied to preserve features such as eye-blinking which can be observed in daily human motion~\cite{richard2021meshtalk,Xing_2023_CVPR}.

Simply regressing the speech to the expression may sacrifice the emotional factor of the face:  
Karras et al.~\cite{karras2017audio} introduce the emotion vector to which the emotion parameters are encoded during training. 
Peng et al.~\cite{Peng_2023_ICCV}  disentangle the emotion from the data  using an emotion encoder. 
Daněček et al.~\cite{EMOTE} learn emotional facial expressions using a temporal VAE structure; in addition to the speech, the emotion label is given as an input to control to produce rich expressions of the characters.  
Further, they improve the quality of lip synchronization by using an image projection loss when learning the motion from videos.  
Fan et al.~\cite{fan2022joint} adds the language features to the input for controlling the emotion of the character.   
Several methods make use of diffusion models to improve the variation of the expressions during speech~\cite{stan2023facediffuser}.
Most existing works are focuses on the motion synthesis from a single speaker. 

Producing body motion from speech has also been explored in computer graphics, computer vision and machine learning areas.  
The issue of the problem of speech to gesture is the weak correlation between the two signals: the person can move their body in very different ways during their speech. 
Early approaches directly regress the speech features to the body motion, which suffer from smoothing artifacts and lack of contextual motion~\cite{ginosar2019learning  }.  To reduce such artifacts, methods based on motion matching~\cite{Habibie_gesturematching22}, generative models such as normalizing flows~\cite{DBLP:journals/cgf/AlexandersonHKB20,valle2021transflower}, variational transformers~\cite{pang2023bodyformer} and diffusion models~\cite{alexanderson2023listen} are developed.  
Ghorban et al.~\cite{ghorbani2023zeroeggs} provide template motion to control the style while Ao et al.~\cite{ao2023gesturediffuclip} provide prompt as an additional input.

Very recently, methods to synthesize both the face and body motions via the speech are developed~\cite{yi2023generating,liu2023emage,ng2024audio2photoreal}.
SHOW~\cite{yi2023generating} trains separate VQ models for each mod, body and fingers to simultaneously control them through speech. 
EMAGE~\cite{liu2023emage} prepares transformer-based controllers for different parts of the body to produce motion that includes global translation. Ng et al.~\cite{ng2024audio2photoreal} propose a VQ/diffusion-based frame-work to individually train the face and body motions which are unified to output photo-realistic face and body motion.  

Also, there has been an increasing interest in producing motions based on conversations instead of just monologue speech
~\cite{ng2021body2hands,ng2022learning,Ng_2023_ICCV, pang2023bodyformer,ng2024audio2photoreal}: in such situations, the body will move not only move in synch to speaker's spontaneous speech but also in response to the other person's speech.  It has been shown motions such as nodding and denial behaviours  of the listener's motion can be automatically produced from the speaker's speech.  On the other hand, previous methods regress the speech but not other modalities such as the distance between characters, facial expressions and body poses to the character, which should also affect their motions.

In our research, we also propose a multi-modal model where the motions of two characters are simultaneously generated through the conversation. Compared to previous models, the motion is not only controlled by the speech but also by the face, body motion and the relative position/orientation with the other speaker. We show such multi-modal inputs are needed for producing realistic response of the character. 

\subsection{Datasets}
The quantity and quality of data are critical for the performance of both types of methods. Many audio-driven human motion generation datasets exist and they mainly differ in the number of persons, types of audio input and motions, and the scenarios the actors are supposed to perform. We list and compare the most popular existing datasets in Table \ref{tab:dataset_comp}.

We emphasis three key differences between our dataset and the previous ones. Firstly, our dataset is multi-modal and contains body motions and 3D facial meshes of two persons, and also their speeches. In contrast the existing datasets which either only consider one single person's motion and monologue speech, or the motions and speeches of two persons without the facial expressions. Secondly, in our dataset for every motion the pair of actors are assigned a unique relationship and two roles. They are expected to act out a scenario with each other. Thus the motion sequences in our dataset exhibit way higher coherence than the ones in other datasets. Thirdly, the persons in the previous dataset are mostly static or move within a small range of space, while the actors in our dataset are much more dynamic and there can be bigger variation as to the trajectories. 

\begin{table}[]
    \centering
    \begin{tabular}{c|ccccccc}
        \toprule
        \#Actor & Name & \#Persons & Type & Face & \#Sequence & Avg.Len. & Duration\\        
         \midrule
       \multirow{4}{*}{1} & TalkSHOW\cite{yi2023generating} & 4 & Monologue & \checkmark & 9720 & 10.0s & 27.0h\\   

       & ZeroEGGS\cite{ghorbani2023zeroeggs} & 1 & Monologue &  \ding{55} & 67 & 2.0min & 2.3h \\

       & Motorica\cite{alexanderson2023listen} & 5 & Dancing & \ding{55} & 244 & 1.7min & 7.0h\\        

       & BEAT\cite{liu2022beat} & 30 & Monologue & \checkmark & 2508 & 1.8min & 76.0h\\
       \hline 
       \multirow{4}{*}{2}& Ng et al.\cite{ng2024audio2photoreal} & 4 & Conver. & \checkmark & 4 & 2.0h & 8.0h \\        
        & Inter-X \cite{xu2023inter} & 89 & Interaction & \ding{55} & 11388 & 6.0s & 18.8h \\
       & Hi4D\cite{yin2023hi4d} & 20 & 
       Interaction & \checkmark & 100 & 3.6s & 6.1min \\    
       & Ours & 7 & Conver. &  \checkmark & 241 & 60s & 8.3h \\
        \bottomrule
    \end{tabular}
    \caption{Comparison between our new dataset and existing datasets. Ours is the first to simultaneously capture audios, body motions, and facial meshes of two actors. }
    \label{tab:dataset_comp}
\end{table}

\section{Methodology}
\begin{figure}
    \centering
    \includegraphics[width=1.0\linewidth]{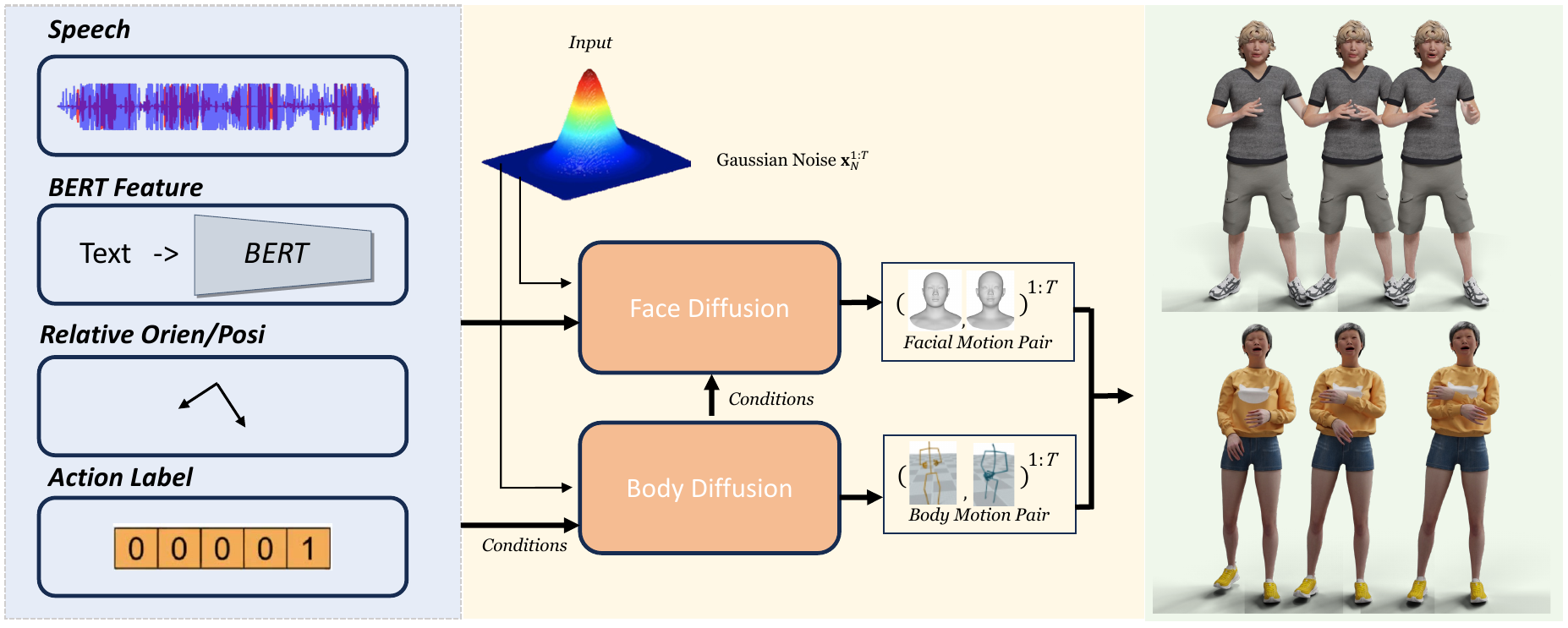}
    \caption{Our method's pipeline begins by taking the raw audio signals from two individuals. It then incorporates a suite of conditions, including BERT~\cite{devlin2018bert} features, relative orientation and position data, as well as action labels. Utilizing these inputs, our approach employs two distinct diffusion models in tandem to generate lifelike and varied facial and body animations. The result of our algorithm is the simultaneous production of intricate 3D facial meshes and comprehensive global body movements for both subjects.}
    \label{fig:pipeline}
\end{figure}

Similar with previous works \cite{alexanderson2023listen}, we propose to adopt the recently proposed Diffusion Models \cite{sohl2015deep, ho2020denoising, song2019generative} to model the mapping from the audio inputs to the corresponding body motions and facial expressions for the two persons simultaneously. In contrast to most previous works 
that model each person individually, we train our model to control two persons simultaneously. The framework of our system is shown in Fig. \ref{fig:pipeline}.
In this section, 
we first describe about the details of our large two person interaction dataset for training our model, 
and then describe the details of our model. 

\subsection{Dataset}
\subsubsection{Content}

Due to the multi-modal nature of our proposed model, a unique dataset that comprises of facial, body, speech, and varied relationships and emotions was required. To this end, we constructed and utilized a rich, two-person dataset, comprising of 241 captured scenarios, across 25 inter-personal relationships and 26 emotions.

The unique dataset constructed for our method stands out due to its comprehensive coverage of relationships and emotions, which spans over settings such as family, work, friends, and school. Our selection covers distinct relationships including, but not limited to, siblings, parent-child, doctor-patient, and teacher-student interactions. The emotional spectrum of our dataset references the work of Cowen et al. \cite{doi:10.1073/pnas.1702247114} and incorporates a modified subset of emotions that exist in distinct categories, being bridged by continuous gradients. These emotions include joy, adoration, interest, boredom, anxiety, among other, totaling 26 in all.

To facilitate the coverage of each pair of relationship and emotion, we devised a one-sentence character setup and a scenario description. Table \ref{tab:scenarios} includes a small sample of scenarios captured in our dataset.

\begin{table}
    \centering
    \begin{tabular}{p{2cm}p{2cm}p{3.5cm}p{3.5cm}}
        \toprule
       Relationship & Emotion & Character Setup & Scenario Description \\
        \midrule
        Siblings & Joy & An older brother, 20 years of age, and his teenage sister, who is 16 years old. & They both unexpectedly receive an acceptance letter from their respective dream institutions. \\
        Co-worker & Romance & Two co-workers, a project manager in her 30s and a software engineer in his 40s who have known each other for several years. & During a late-night working session at the empty office, they confess their feelings for each other. \\
         \bottomrule
    \end{tabular}
    \caption{Excerpt of scenarios in our dataset}
    \label{tab:scenarios}
\end{table}

Our dataset also ensures diversity by including 8 male and female actors from varying age groups and nationalities. For every scenario, a pair of male and female actors perform impromptu for a duration of one minute or more, without any provided script apart from the given Character Setup and Scenario Description prompts. Two chairs are available to them as a prop, which they can freely arrange or use to simulate various real-life situations, such as the front seats of a vehicle, a classroom, or a doctor's office.

\subsubsection{Data Capture}
The capture system consists of several components either worn by the actor or stationed around the capture space. To capture both actor's body motions, a 28-camera VICON optical motion capture system was positioned around the 5m x 5m space, with each camera located in one of three elevations. 53 body markers and 20 finger markers are placed on each individual actor, and a Range of Motion exercise is performed for actor calibration.

For the face, the front depth sensor and camera of an iPhone is utilized to capture the facial movements of the actor. Prior to the capture, a mesh template of each actor's front face is first registered by asking them to rotate their head while facing the camera. During the capture, an actor will wear a head rig that consists of the iPhone, two microphones (one active and one backup), and a power bank that also acts as a counterweight.

As the motion and face capture systems are separate, they need to be temporally synced for accuracy. To achieve this, a wireless timecode generator is used to broadcast the clock signal of the VICON system to individual iPhones, syncing the timecode at the beginning of a shoot. Then, before and after every captured scenario, a capture script is triggered to synchronously start and stop the recording of both systems at a given timecode. This automatically ensures frame-level accuracy while requiring less post-processing work to manually sync the two sources together. During the capture, the live audio and face video of the two actors are streamed to the control station for real-time monitoring, which ensures data quality.

\begin{figure}
    \centering
    \includegraphics[width=0.9\linewidth]{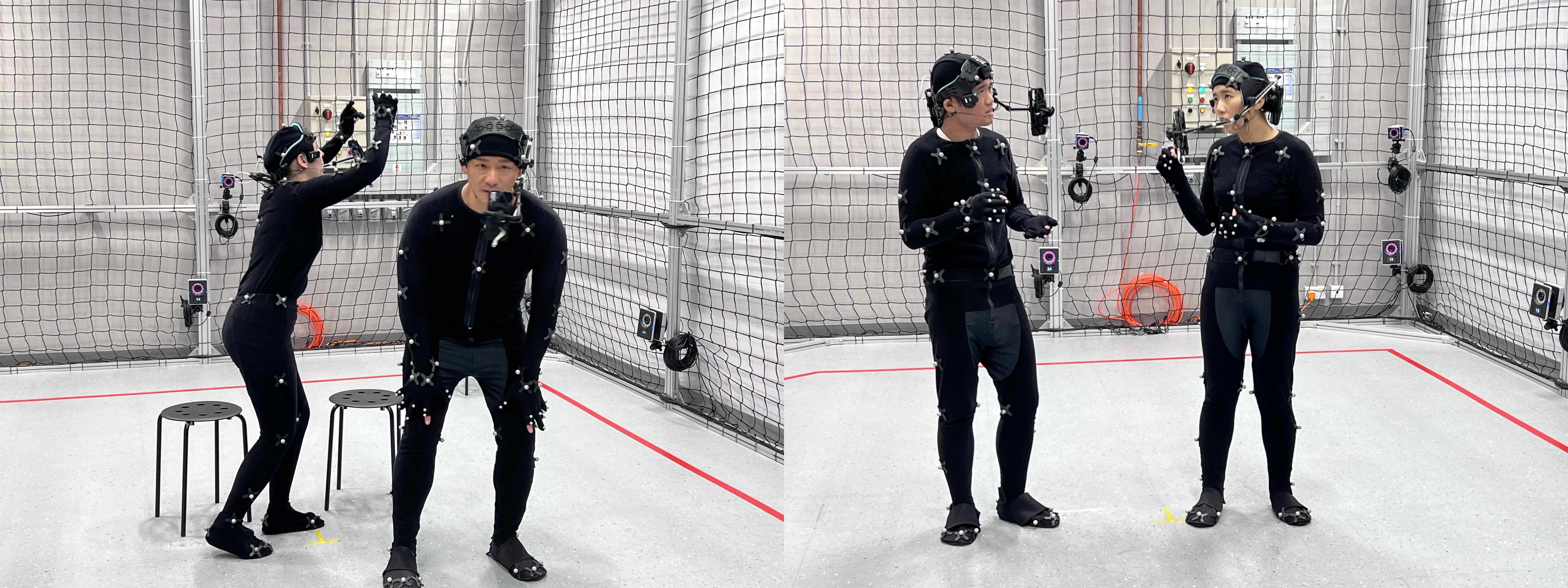}
    \caption{Actors during performance, showing the body and face capture setup}
    \label{fig:capture_actors}
\end{figure}

\subsection{Pre-processing}
Though it is possible to directly treat raw audio signal as the input, previous methods always firstly extract features from the audio. We follow the common practice here and also conduct pre-processing on the audios from both actors in separation, then combine the outputs together as the new representation. More specifically, as is done in BodyFormer \cite{pang2023bodyformer}, we consider both low-level audio features and high-level semantic textural features. We choose the 27-d mel-spectrogram as the proxy for low-level features, which is expected to encode the speech attributes for each actor. The BERT representation \cite{devlin2018bert} is used for the high-level feature. In the end, for every frame, there is a 59-d vector representing the audio input. We also compute the action mode for every frame, and assign a unique label among SIT, WALK and STAND for each frame. The one-hot encoding of this action label extends the input to 62-d.

As the output, the body motion can be specified in different formats, like flow of raw body vertices, a sequence of SMPL pose parameters \cite{loper2023smpl}, or a trajectory of 3d skeletons \cite{monszpart2019imapper, savva2016pigs}. In this work we use a list of the relative orientation of each joint with respect to its parent joint as the motion representation. Note that this is the standard format used in the ubiquitous BVH file format, and can easily be converted into 3D joints or SMPL parameters. We further reduce the complexity of the body motion by transforming the orientation of each joint into the local frame of the same joint in the previous frame. This way the motion representation is invariant to global body orientation, and also has a smaller rotation angle range. Exponential map \cite{grassia1998practical}  is adopted as the representation for the local rotation, as the effectiveness of this kind of representation has been verified in the previous works \cite{alexanderson2023listen}.

\subsection{Interactive Motion Estimation}
We target at estimating the interactive motion between two actors driven by their dialogue. Similar exploration has been made by MDM \cite{tevet2022human} and Inter-X \cite{xu2023inter}. While they focus on generating expressive and realistic human motions from textural descriptions, in this work we address the multi-modality and interactive nature of human motion estimation from the conversation and their spatial relations. 

The model we use to regress the humans' motions from their speech is built on top of the model proposed in \cite{alexanderson2023listen}, where a variant of Conformers \cite{zhang2022music} is combined with diffusion models. Different from the initial work where only one person's motion is estimated from the audio signal, we adapt the base model to estimate the motion of two persons from their speech signals, BERT features of their speech contents, one-hot action labels (STAND,SIT,WALK) and their initial relative position and orientation.   
In this way the contents of their speech and the global interaction between the two persons is considered by our work. Formally, given a dataset of $N$ evenly segmented input features and the corresponding body motions:
\begin{equation}
D = \{(X_1^1, Y_1^1, X_1^2, Y_1^2), (X_2^1, Y_2^1, X_2^2, Y_2^2) ... (X_N^1, Y_N^1, X_N^2, Y_N^2)\},
\end{equation}
where $X_i^j$ and $Y_i^j$ with $1 <= i <= N$ and $j \in {1,2}$ refer to the $j$-th person's input features and body motion for $i$-th motion segment, we firstly concatenate the features and output body motions for two persons, thus having a new dataset: 
\begin{equation}
    D' = \{(X_1, Y_1), (X_2, Y_2) ... (X_N, Y_N)\}.
\end{equation}

We want diffusion models to model the mapping from input feature to the body motions and facial expressions in separation. Take the body diffusion model as example. The diffusion process is formulated as a Markov diffusion process, $\{Y_t\}_{t=0}^T$, where $T$ is total number of diffusion steps, $Y_T$ is randomly sampled from normal distribution $\mathcal{N}(0, I)$ and $Y_0$ is the drawn real body motions from the processed dataset $D'$. The transition from $Y_{t-1}$ to $Y_t$ is formulated as:
\begin{equation}
    q(X_t | X_{t-1}) = \mathcal{N}(\sqrt{\alpha_t x_{t-1}, (1-\alpha_t)I})
\end{equation}
where $\alpha_t \in (0,1)$ are fixed hyper-parameters. We follow the practice of the method in \cite{alexanderson2023listen, tevet2022human} to use the simple training loss
\begin{equation}
E_{Y_0 \sim q(Y_0 | X), t \sim [1, T]} [ \| Y_0 - G(Y_t, t, X)\|_2^2 ]
\end{equation}
to train the Conformers based diffusion models to learn the denoising process, thus in the end our method can generate realistic body motions from the initial randomly drawn noise given the speeches.

\subsection{Body-conditioned Facial Expression Estimation}
We propose a model to simultaneously regress the facial expressions of two actors driven by dialogue, as well as a body feature. We note that a person's facial expressions and eye movement can differ substantially depending on whether they are currently facing the other person, i.e. the direction of their gaze, the context of the dialogue, as well as the spoken content and facial expression of the other person. Hence, the incorporation of such information during the generation of facial motions leads to more accurate and naturalistic facial animation results.

The model for regressing both person's facial animations is built on top of the DiffSpeaker model proposed in \cite{ma2024diffspeaker}, where the auto-regressive FaceFormer \cite{fan2022faceformer} architecture with alignment and attention bias used for frame-by-frame animation generation is instead adapted for learning a denoising task within a diffusion model. We adapt the model to support two-person audio and vertex sequences as input, and the pose information as condition.

Specifically, for two persons $A$ and $B$, we aim to produce the concatenated facial motion $v_{A^\frown B}^{1:T}$ of time duration $T$ given the synchronized two-person dialogue audio $a_A^{1:T}$ and $a_B^{1:T}$, the speaker style $s_A$ and $s_B$, as well as a one-hot embedding $p \in \mathbb{R}^2$ denoting whether the two persons are facing each other. The entire network $\textbf{F}$ can be formulated as:

\begin{equation}
    \hat{v}_{A^\frown B} = \textbf{F}(a_A, a_B, s_A, s_B, p).
\end{equation}

During training, the two person's vertex sequences $v_A^{1:T}$ and $v_B^{1:T}$ are concatenated on the vertex dimension, giving $v_{A^\frown B}^{1:T}$. This sequence is then encoded into 512-d, giving the motion encoding $x_{A^\frown B}$. For audio, $a_A^{1:T}$ and $a_B^{1:T}$ are first seperately encoded using Wav2Vec2 to produce vectors $w_A^{1:T}$ and $w_B^{1:T}$. Afterwards, the vectors are concatenated and passed through an encoder to produce the final audio representation $e_a$. Similarly for conditions, for the speaking style $s_A$ and $s_B$, the one-hot embedding $p$, and the diffusion step information $n$, they are passed through their respective condition encoders to produce $e_s$, $e_p$, and $e_n$. The diffusion denoiser $\textbf{D}$ can therefore be formulated for training as:
\begin{equation}
    \hat{x}_{A^\frown B}^0 = \textbf{D}(x_{A^\frown B}^n, e_a, e_p, e_n)
\end{equation}
where $x^n$ denotes the latent facial animation $x$ added with Gaussian noise, with $x^0$ denoting the latent animation without noise added, and $x^N$ denoting pure noise.

For the cross-attention mechanism in the denoiser Transformer architecture, we extend the \textit{Biased Conditional Attention} in Diffspeaker which utilizes Prefix Tuning as proposed by \cite{pfeiffer2020AdapterHub} to include the pose encoding $e_p$ as an extra condition. The attention score ($\textbf{S}$) formula is hence modified as:

\begin{equation}
    \textbf{S} = Softmax(Q{[e_s, e_p, e_n, K]^T} + B)
\end{equation}
where $Q$, $K$, $B$ denotes the query, key, and bias terms of the attention score respectively. 

\section{Experiments}
We conduct statistical analysis on the newly captured dataset, and also train and validate audio-driven interactive human motions and facial expressions on the new dataset. Once trained, our algorithm based on diffusion models can directly yield global body motion and facial animations for both actors, 
with the audio signals, the BERT features of the speech and the relation position/orientation of the two. 

\begin{table}[h]
    \centering
    \resizebox{\columnwidth}{!}{
    \begin{tabular}{c | c  c | c c c c c}
        \toprule
        Type & Frames & Percentage & Upper.LArm & Upper.RArm & Lower.LLeg & Lower.RLeg & Root \\
        \midrule
        Facing & 173031 & 31.78 & 21.31 & 13.95 & 22.86 & 21.42 & 5.95 \\
        Not-Facing & 371565 & 68.22 & 18.40 & 15.43 & 25.98  & 25.77 & 6.15 \\ 
        \hline
        Co-worker& 53641 & 9.82 & 17.92 & 16.41 & 22.19 & 19.37 & 6.49 \\
        Romantic Partner & 76074 & 13.97 & 13.00 & 16.66 & 28.87 & 28.60 & 2.23 \\
        Waiter / Diner & 53074 & 9.75 & 16.67 & 13.18 & 23.56 & 24.59 & 7.45 \\
        Doctor / Patient & 58020 & 10.65 & 8.37 & 10.60 & 5.68 & 7.59 & 3.08 \\       
        \bottomrule
    \end{tabular}
    }
    \caption{(
    Left) Number of frames in Facing and Not-Facing poses,
    and different relationship with percentage; (Right) Standard deviation of rotation angles for different joints with respect to facing/non-facing as well as four different relationships. The unit is degree. It is easy to see the range of rotation angles is much smaller when the two persons are facing each other except for the upper left arm.  The doctor/patient relationships also result in smaller range of motion. 
    }
    \label{tab:angle_std}
\end{table}

\subsection{Dataset Analysis}
Here we do an analysis of the capture data to show there is a strong correlation between the relative position of the two actors (facing or non-facing), the relationship between the two, as well as the emotions and the corresponding facial/body motion, which justifies the multi-modal architecture that we adopt in our system.    

\paragraph{Body} We define an actor to be in the Facing Pose if for both actors, the other actor's head exist within a 60\textdegree arc of the actor's central vision. This is because within this angle range, one can rotate their eyeball to naturally converse or interact with the other actor. Table \ref{tab:angle_std} shows the distribution of Facing and Non-Facing poses in the dataset - it can be observed that overall the diversity of motion is smaller when the actors are facing each other, could be due to the smaller personal space and the restriction to face the other actor.  This shows that it is essential to condition the body motion on the relative position/orientation of the two. 

Additionally, it can be observed that the actors' pose variance is reduced for the doctor/patient relationship compared to the others, which could be due to the two actors mainly sitting and talking about the diagnosis.  
This also shows that the contents of the speech can greatly affect the body motion during the interaction.

\begin{figure}[h]
    \centering    \includegraphics[width=1\textwidth]{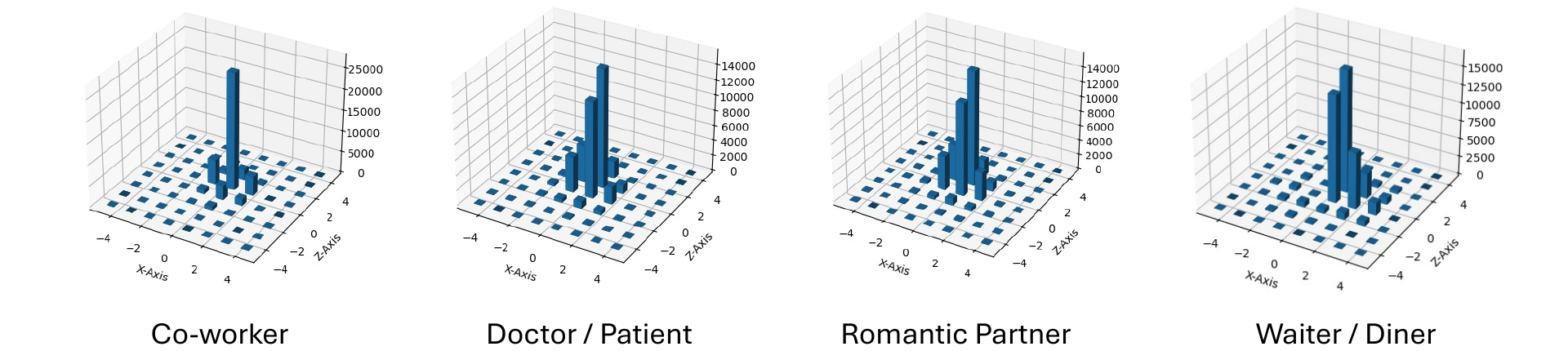}
    \caption{Histograms of the second actor's root position on the ground plane (XZ plane) relative to the first actor, for 4 different kinds of relationships. The unit is meter. Note how the relative position of the second actor with respect to the first actor varies depending on the relationship of the actors. }
    \label{fig:body_variance}
\end{figure}

Finally, we show the distribution of the relative position/orientation of the two actors in Figure \ref{fig:body_variance}.

\begin{figure}
    \centering
    \includegraphics[width=1\textwidth]{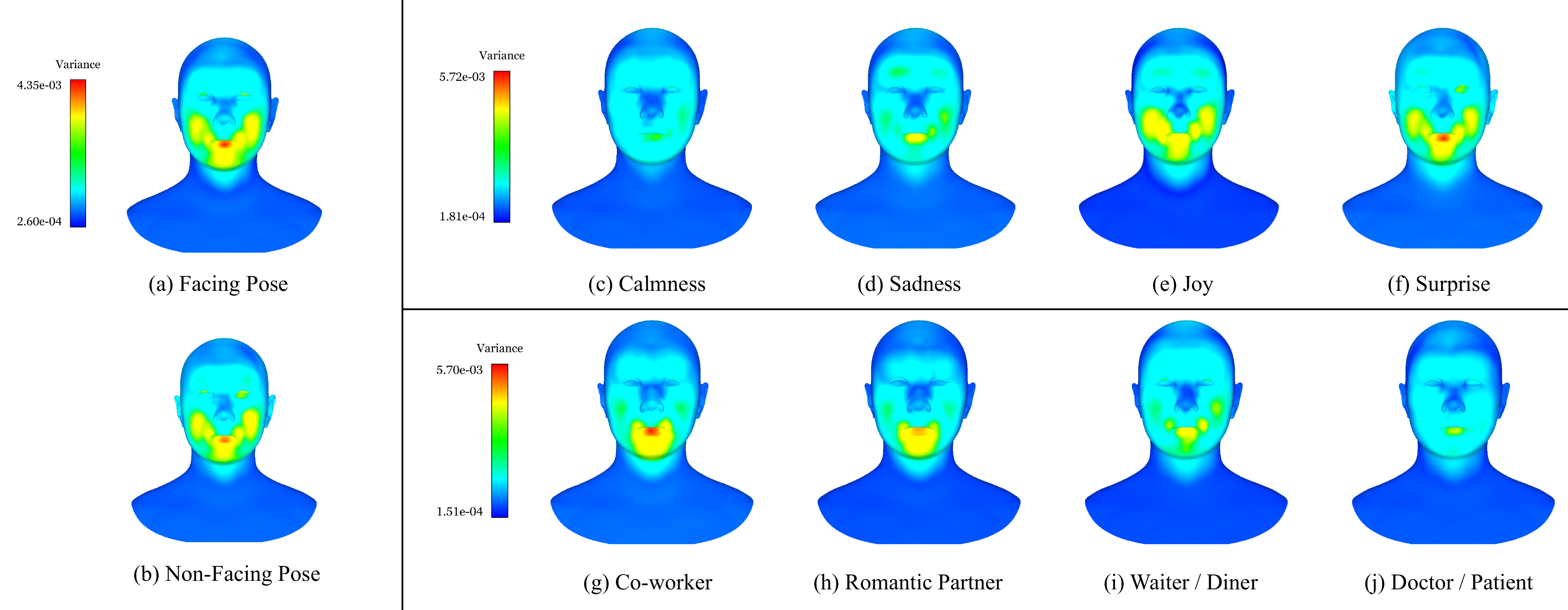}
    \caption{The heat map of face animation variance between different facing directions(Left), emotions(Right Top), and relationships (Right Bottom). }
    \label{fig:face_variance}
\end{figure}

\paragraph{Face} We visualize the diversity of facial animation based on different speaking poses, emotions, and relationships in Figure~\ref{fig:face_variance}, focusing on the distribution of the captured facial performance by visualization of the variance.

From the left, we see that the area around the eye in the non-facing pose has a higher variance compared to that in the facing pose. This likely indicates that actors, when not facing each other, tend to look around more, leading to the increased variance. Furthermore, a comparison of the lip area indicates a higher variance for the facing pose, demonstrating greater lip movement when actors are facing each other.
Emotion and relationship also greatly influence facial animation.
The variance of the lip area for surprise is the highest among all emotions, indicating a tendency for actors to open their mouths widely for said portrayal; 
Romantic partners exhibit a larger variance due to the stronger emotions portrayed. Furthermore, the doctor may exhibit lower facial variance due to the depiction of professionalism.

\subsection{Audio-driven Motion Generation}
We now provide a quantitative evaluation of the full body motion generated by our system.  We first provide the metrics of evaluation and then the comparison of our system with state-of-the-art architectures.  

\paragraph{Metrics}
Following prior works ~\cite{alexanderson2023listen, ng2024audio2photoreal}, we employ a suite of metrics that capture both the realism and diversity of generated motions. 
1. The Fréchet Inception Distance (\textbf{$FID$}) is a widely accepted metric in generative tasks, 
quantifying the distributional distance between ground truth and generated data, with various features extracted to represent different aspects of the generated motion. 
The $FID_{g}$ focuses on "geometric" realism by measuring this distance for static poses of two persons, adjusted for orientation to face the positive x-axis. 
The $FID_{k}$ metric extends the evaluation to "kinetic" realism by collaborating with a pre-trained feature extractor. This extractor encodes the pose sequences of a single individual into a latent representation, facilitating the measurement of motion quality in a dynamic context.
For evaluating "two-person relationship" realism, $FID_{r}$ calculates the joint distances map between two individuals. 
2. We also assess the diversity $DIV$ by computing the average pairwise distance of generated samples. 
3. Lastly, the $Foot.Slid$ metric gauges physical plausibility by quantifying the extent of foot sliding or skidding, with lower values indicating more realistic foot behavior.

In Table~\ref{tab:quantitative_body}, we conducted an evaluation of body motion generation. 
In comparison to the baseline LDA model, which also incorporates audio input, our method exhibited lower performance for the single person's motion, in terms of the dynamic and physical realism.
This discrepancy stems from a more complex conditional space in  task setting, that our model is designed to generate motion that is cognizant of the presence of another individual.
Despite this, our method demonstrated a superior capability in capturing the nuances of two-person interactions. 
It produces more convincing static poses and maintains better relative joint distances. 
Furthermore, the diversity of interactions between two individuals is enhanced through the incorporation of relative offsets as an additional input, enriching the overall interaction modeling.

\vspace{8mm}
\noindent %
\begin{minipage}[t]{0.55\textwidth} 
    \centering 
    \resizebox{\textwidth}{!}{
    \begin{tabular}{c|ccccc}
        \toprule
        Method & $FID_{g}\downarrow$ & $FID_{k}\downarrow$ & $FID_{r}\downarrow$ & $DIV\uparrow$ & $Foot.Slid\downarrow$ \\
        \midrule
        LDA~\cite{alexanderson2023listen} & 0.318 & \textbf{0.445} & 3.831  & 0.460 & \textbf{0.0033} \\   
        W/o offset & 0.463 & 0.965 & 5.088 & 0.550  & 0.0043 \\
        Ours & \textbf{0.280}  & 0.876 & \textbf{1.421} & \textbf{0.730}  & 0.0042 \\
        \bottomrule
    \end{tabular}
    }
    \captionof{table}{Quantitative evaluation of the body motion generation. All model is trained on our dataset and we highlight the top.}
    \label{tab:quantitative_body}
\end{minipage} \hfill
\begin{minipage}[t]{0.38\textwidth}
    \centering 
    \resizebox{\textwidth}{!}{
    \begin{tabular}{ccc}
        \toprule
        Method & $LVE \downarrow$ & $FDD \downarrow$ \\
        \midrule
        FaceFormer & 4.1468e-05 & 9.0007e-05 \\   

        Ours(Single) & 4.0757e-05 & \textbf{8.4152e-05} \\        

        Ours(Both) & \textbf{4.0278e-05} & 8.6616e-05 \\
        \bottomrule
    \end{tabular}
    }
    \captionof{table}{Quantitative evaluation of the face motion.}
    \label{tab:quantitative_face}
\end{minipage}
\vspace{8mm}

\subsection{Audio-driven Facial Expression Generation}
Next we provide a quantitative evaluation of the face motion generated by our system.  We first provide the metrics of evaluation and then the comparison of our system with state-of-the-art architectures.  

\paragraph{Metrics}
Facial Expression accuracy and fidelity can be measured by considering the lip synchronization and upper face variations separately. The following metrics as proposed in ~\cite{richard2021meshtalk, xing2023codetalker} are used as a benchmark:
Lip Vertex Error ($LVE$), measured as the averaged maximum $l_2$ error among all lip vertices across all frames. The lower the value, the more accurate the lip synchronization. This is a common metric used in ~\cite{richard2021meshtalk, fan2022faceformer, ma2024diffspeaker};
Face Dynamics Deviation ($FDD$), proposed by \cite{xing2023codetalker}, is measured as
        \begin{equation}
            FDD(M_{1:T}, \hat{M}_{1:T}) = \dfrac{\sum_{v\in \mathcal{S}_U} (dyn(M_{1:T}^v) - dyn(\hat{M}_{1:T}^v))}{\mid \mathcal{S}_U \mid}
        \end{equation}
where the temporal standard deviation of the $l_2$ element-wise norm is calculated. This metric aims to quantify the variation of upper facial dynamics by means of comparing deviation over time.

We trained the baseline model using our dataset and evaluated it against our approach using a test split of 31 two-person sequences that the models had not encountered during training. The results, as displayed in Table \ref{tab:quantitative_face}, indicate that our method surpassed the baseline in delivering superior results, particularly in lip synchronization and facial dynamics.

\subsection{Qualitative Comparison}

\begin{figure}[t]
    \centering
    \begin{subfigure}[b]{1.\linewidth}
        \includegraphics[width=1.0\linewidth]{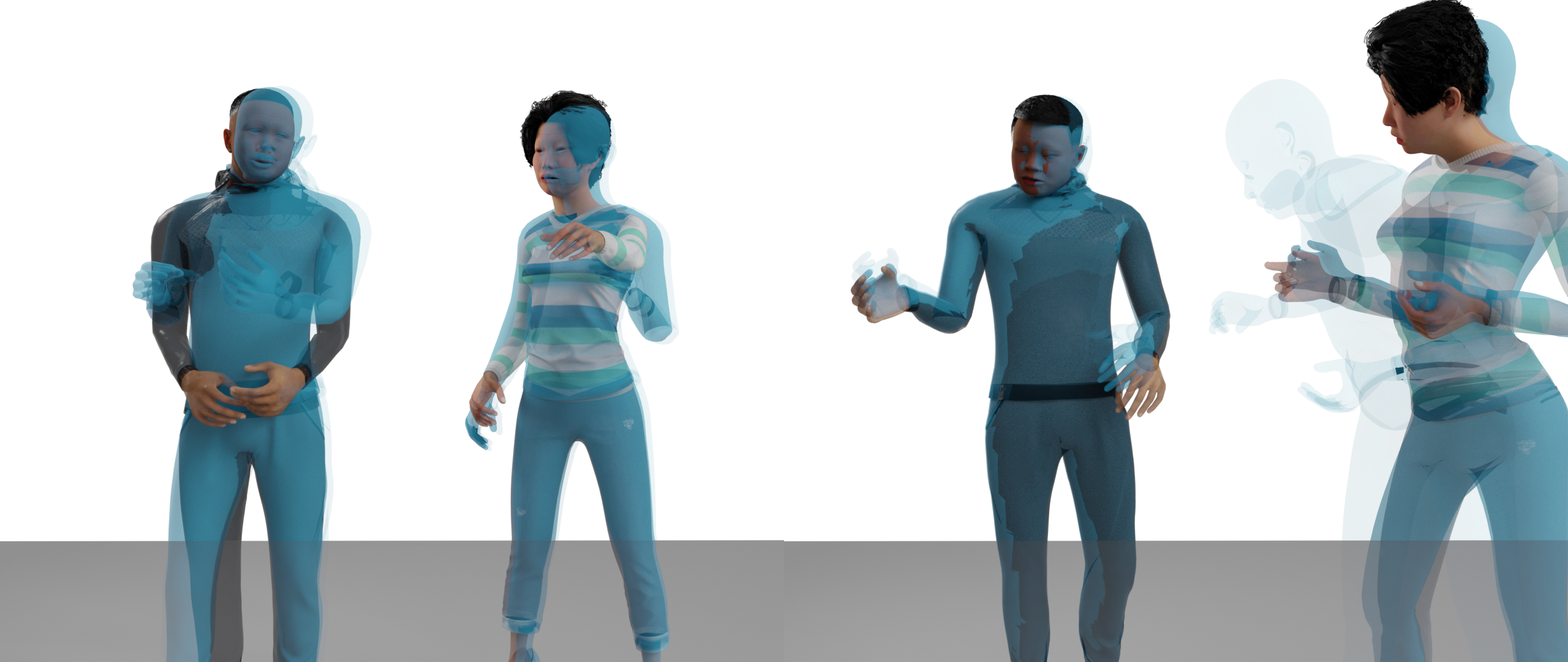}
    \end{subfigure}            
    
    \begin{subfigure}[b]{1.\linewidth}
        \centering
        \includegraphics[width=0.3265\linewidth]{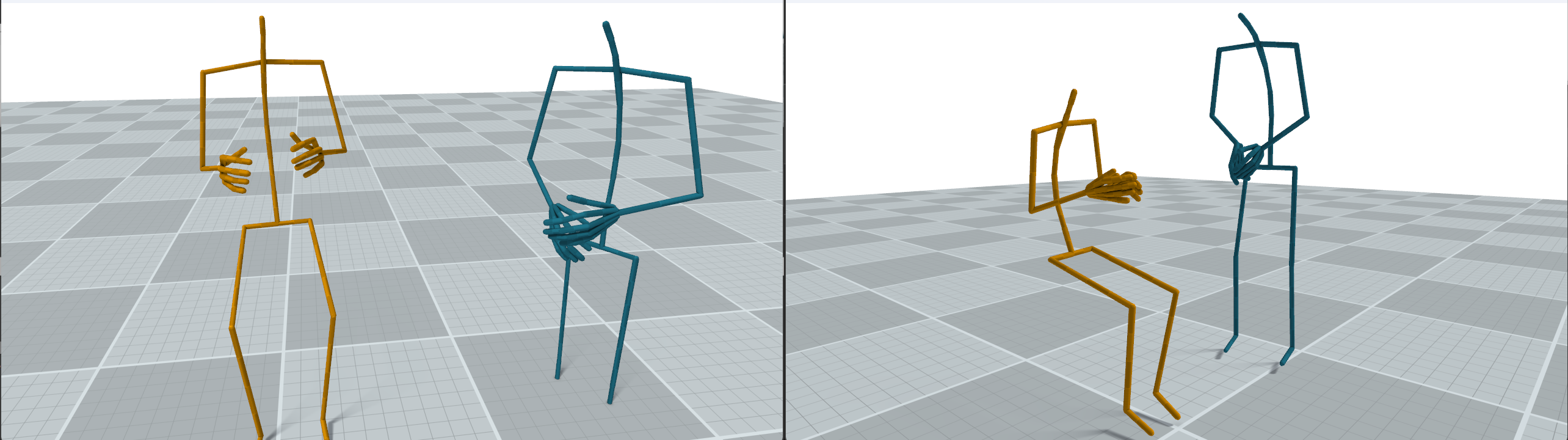}    
        \includegraphics[width=0.3265\linewidth]{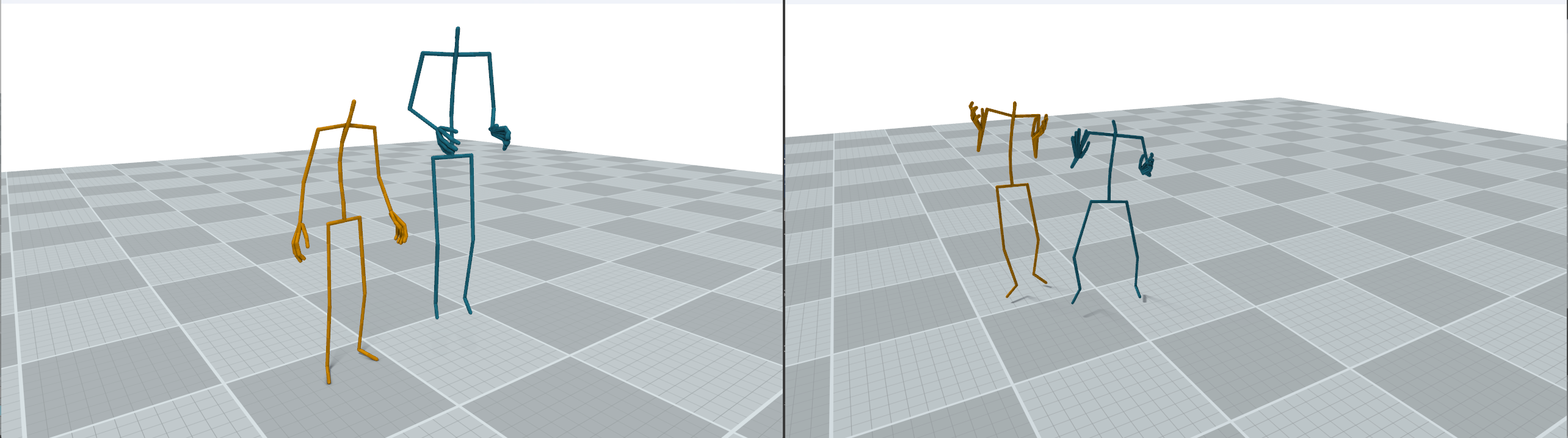}   
        \includegraphics[width=0.3265\linewidth]{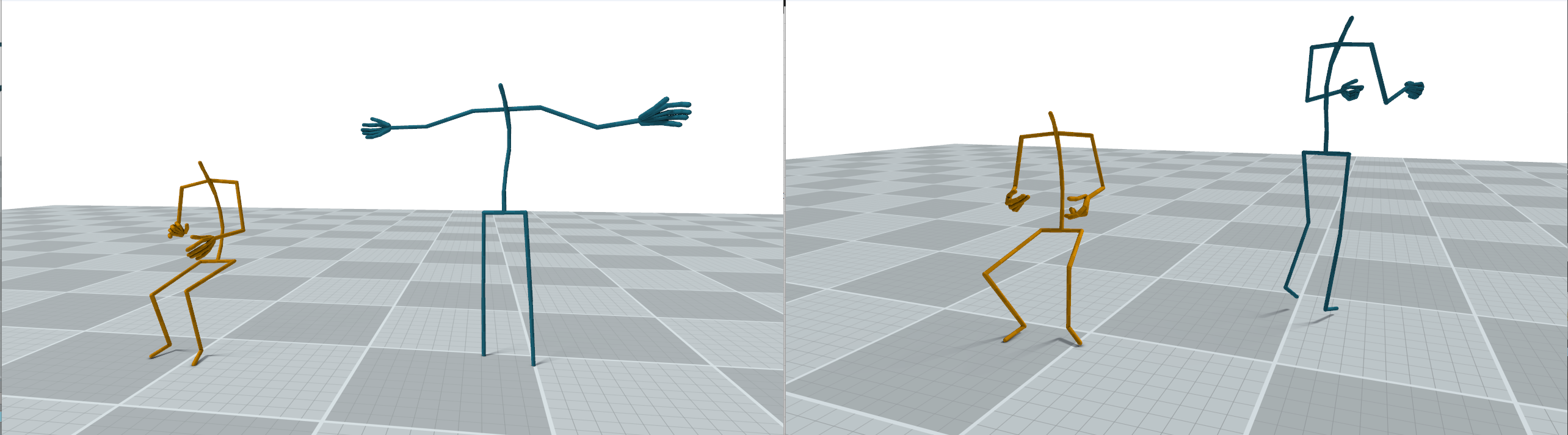}         
    \end{subfigure}            

    \caption{Visual comparison between the predictions of our method with the ground-truth motions. For each pair of images, the left one is our predicted result, and the right one is the Ground-Truth. (Top) \textit{Neighbors reminiscing about their past.} Conditioning on the same audio input, we generate plausible full-body motions different from GT. (Bottom) Left two pairs are from the Waiter/Customer relationship, and in the second pair the waiter shows emotion of fear. The third pair is of the relationship of Co-worker. The scenario goes like this: \textit{the older coworker is eating a disgustingly smelly lunch at his desk, and the unfazed reactions of those around him shocks and disgusts the new hire}.}
    \label{fig:compare_955}
\end{figure}

To assess the visual realism and precision of our technique, we rendered both the predicted motions from our method and the ground truth motions corresponding to the same speech, as illustrated in Fig. \ref{fig:compare_955}.

Thanks to our extensive and varied dataset, coupled with a robust baseline system, our method is able to generate motions that are not only convincing but also exhibit diversity when given the same input.
Our approach also offers the flexibility to tailor the generated motions by applying different contextual conditions, such as the nature of the relationship, action labels, and emotional states.
Please refer to our supplementary video for more details. 

\section{Conclusions}
In this paper we target modelling the ubiquitous interactive activities between two persons happening in various daily scenarios. To this end we firstly captured a medium-sized multi-modal dataset which contains audio, body motion, and facial expressions. Altogether 4 pairs of professional actors perform 241 motions covering diverse relationships and emotional states, totalling in around 6 hours of data. We conduct detailed statistical analysis on the new dataset, and show that whether two persons are facing each other has a big influence on their motions. We also propose two diffusion model based algorithms to predict body motions and facial expressions of the persons from their audio signals alone. Experiments verify the effectiveness of the proposed algorithms.
\clearpage

\bibliographystyle{splncs04}
\bibliography{main}
\end{document}